# From Simulations to Reality: Enhancing Multi-Robot Exploration for Urban Search and Rescue

**Gautam Siddharth Kashyap[1], Deepkashi Mahajan[2], Orchid Chetia Phukan[3], Ankit Kumar[4], Alexander E.I. Brownlee[5], Jiechao Gao[6]**
[1]officialgautamgsk.gsk@gmail.com, [2]mahajandeepakshi03@gmail.com, [3]orchidp@iiitd.ac.in,
[4]ankitk@iiitd.ac.in, [5]sbr@cs.stir.ac.uk, [6]jg5ycn@virginia.edu
[1, 3, 4]IIIT Delhi, New Delhi, India
[2]GeeksforGeeks, Noida, India
[5]Division of Computing Science & Mathematics, University of Stirling, UK
[6]University of Virginia, Virginia, USA

*Abstract*—In this study, we present a novel hybrid algorithm, combining Levy Flight (LF) and Particle Swarm Optimization (PSO) (LF-PSO), tailored for efficient multi-robot exploration in unknown environments with limited communication and no global positioning information. The research addresses the growing interest in employing multiple autonomous robots for exploration tasks, particularly in scenarios such as Urban Search and Rescue (USAR) operations. Multiple robots offer advantages like increased task coverage, robustness, flexibility, and scalability. However, existing approaches often make assumptions such as search area, robot positioning, communication restrictions, and target information that may not hold in real-world situations. The hybrid algorithm leverages LF, known for its effectiveness in large space exploration with sparse targets, and incorporates inter-robot repulsion as a social component through PSO. This combination enhances area exploration efficiency. We redefine the local best and global best positions to suit scenarios without continuous target information. Experimental simulations in a controlled environment demonstrate the algorithm's effectiveness, showcasing improved area coverage compared to traditional methods. In the process of refining our approach and testing it in complex, obstacle-rich environments, the presented work holds promise for enhancing multi-robot exploration in scenarios with limited information and communication capabilities.

## I. INTRODUCTION

In recent times, there has been a growing interest in utilizing multiple autonomous robots for various exploration tasks such as [1]–[4]. One such domain where this technology holds significant potential is in Urban Search and Rescue (USAR) operations. These operations often involve navigating hazardous and unpredictable environments, such as post-disaster scenarios in office buildings, where robots can be deployed to safeguard human lives by minimizing exposure to noxious gases and harmful particulate clouds. However, the advantages of employing multiple robots over single robots are manifold. Multiple robots can efficiently distribute tasks, covering larger areas in shorter time frames. They offer robustness, flexibility, and scalability, making them suitable for diverse applications. In scenarios involving extensive explorations, cooperative strategies among robots can reduce path overlaps, significantly enhancing search efficiency. This approach can be extended to heterogeneous groups of robots, such as combining ground vehicles like rovers with Unmanned Aerial Vehicles (UAVs) for searching large areas. Moreover, multi-robot exploration finds relevance in space science, specifically in lunar and planetary exploration tasks [5]–[7].

Maximizing the coverage of a given space using a limited number of robots, especially when there is little to no information available about the terrain, presents a formidable optimization challenge. This problem requires finding an efficient way for a group of robots to explore the entire space, with the overarching goal of maximizing the fraction of the area covered. In addressing this challenge, metaheuristic algorithms have emerged as valuable tools, offering innovative approaches to optimize robot deployment and space exploration [8]–[12]. In the context of these metaheuristics, each member within the population directly represents an individual robot, with the entire search space mirroring the physical space to be explored by the robots. The fitness of each solution corresponds to the proportion of the space successfully covered, making the application of metaheuristics a promising avenue for tackling this complex problem. Their ability to efficiently explore solution spaces and adapt to diverse scenarios makes them a driving force in various domains in which Particle Swarm Optimization (PSO) [13] emerges as a key player in addressing collective robotic search problems. PSO is an adaptive population-based method that combines social and cognitive behaviors, allowing robots to converge on target locations efficiently. Over the years, PSO has found application in numerous multi-robot search operations due to its adaptability and flexibility [14]–[17]. However, despite the adaptions made to PSO for robotic applications, several assumptions have persisted in previous studies. These assumptions pertain to aspects such as initial information about the search area, robot positioning, communication restrictions, and target information. For example, some studies operate in known search areas [17], [18], while others deal with unknown environments

[19], [20], some with limited boundary information [21], and others in the early stages of development [22]. This paper addresses these common assumptions in multi-robot PSO-based exploration and presents a novel algorithm designed to work in environments devoid of initial information about the environment's size and obstacles. The algorithm is also tailored to operate without global positioning information, considering limited communication capabilities and a lack of information about the target's position. The research contributions of this study are two fold:

**1. Development of a Hybrid Algorithm:** We propose a hybrid algorithm that combines Levy Flight [23] (LF) behavior as a cognitive term and inter-robot repulsion as a social component within the framework of PSO. This novel approach caters to efficient coordinated searching in constrained conditions, such as environments with no initial information on size, obstacles, or global positioning.

**2. Preliminary Simulation Results:** We conduct preliminary simulations to test the performance of our hybrid algorithm in comparison to the basic LF method for multi-robot scenarios. Our results show promising outcomes, with our algorithm outperforming traditional methods in terms of exploration efficiency.

In the subsequent sections of this paper, we delve into related works in Section II, explain the proposed hybrid algorithm's methodology in Section III, discuss obstacle avoidance policies, and present our preliminary implementation results in Section IV. We conclude by outlining our future plans for refining and expanding this research in Section V.

## II. RELATED WORKS

PSO stands as an adaptive population-based technique, where particle behavior results from a blend of social and cognitive behaviors across iterations. It has proven highly effective for addressing collective robotic search problems [24]. PSO empowers robots to follow trajectories, ultimately converging on a target location. Historically, PSO has found extensive utility in various multi-robot search operations due to its versatility and ease of adaptation [25]–[30]. Multiple adaptations to the fundamental PSO framework have emerged to cater specifically to robotic search scenarios, diverging from particle-centric search paradigms. These adaptations, outlined in [31], encompass aspects such as continuous robot movement, movement restrictions, considerations of robot size compared to point-sized PSO particles, the divergence between real-world fitness function calculation and PSO's approach, robot collision with obstacles, and limitations on information sharing. Yet, even with these adaptations, certain assumptions persist to simplify problem statements for diverse scenarios. Four predominant assumptions, recurrent in previous studies on PSO-based multi-robot searching within real-world scenarios, deserve attention. These assumptions pertain to the information regarding the search area, the method of robot positioning within the search space, communication system constraints, and knowledge levels regarding the target's position.

**Search Area Information:** One assumption revolves around the knowledge of the area to be explored. Some studies, such as those in [24], [25], [29], [32], undertake experiments within known search areas. In contrast, studies like [26], [28], [30] delve into unknown search spaces with obstacles, typically operating within confined boundaries and without prior knowledge of obstacle positions. However, research into scenarios with entirely unknown search area dimensions remains nascent [28]. Our study aims to develop an algorithm adaptable to environments devoid of initial size information and capable of achieving optimal search efficiency even in the absence of obstacle data.

**Global Robot Positioning:** Many PSO-based search algorithms assume full knowledge of every robot's position within the environment at all times [26], [28]–[30], [32], [33]. Limited attention has been devoted to scenarios where global positioning information is unavailable [25], [31]. In our current study, the algorithm we've developed can function independently of global robot position data.

**Communication System Restrictions:** Previous PSO-based search algorithms operated with unrestricted communications [24], [29], [32], enabling robots to exchange information freely among themselves and with a central station to determine the global best position. While this centralized coordination enhances efficiency [29], [33], real-world constraints often limit communication capabilities, particularly indoors. Moreover, implementing such capabilities can significantly escalate mission costs. [26], [31] address limited communication scenarios, which our study addresses as well—robots can only communicate within a specific radius, without a central station monitoring their movement.

**Target Information:** The majority of PSO-based robotic search algorithms assume knowledge of the target's position or some form of target-related information. In some cases, the target emits signals detectable by robots from considerable distances [31]. Conversely, [25], [26] assume target detection when robots approach a certain distance, relying on cameras or distance sensors. To tackle this challenge in the absence of target information, alternative algorithms like Random Walk offer flexibility but are often highly inefficient. LF algorithms [34], on the other hand, come into play when exploring vast areas with sparse target distributions.

To address the limitations posed by restricted communication and a lack of information concerning the search space, target position, and robot positions, our work presents a novel multi-robot search strategy—an amalgamation of LF and PSO. This hybrid algorithm incorporates LF behavior as the cognitive component and inter-robot repulsion as the social element within the basic PSO framework. Our research lays out well-defined cooperation strategies for robots, optimizing search efficiency in scenarios where the initial search area size is unknown. Experimental simulations have shown promising results, with multiple robots operating seamlessly in enclosed obstacle-free spaces using this hybrid algorithm. Subsequent research aims to extend these scenarios to larger building environments replete with obstacles and complex boundary conditions.

## III. METHODOLOGY

We introduce a hybrid LF-PSO approach designed for exploration missions in unfamiliar environments. In practical scenarios, robots typically lack initial information about the target's location within the search area. Consequently, they must diligently carry out the exploration task until they approach the target closely enough. When they reach a proximity to the target, a basic camera sensor or infrared (IR) sensor becomes capable of detecting it. Within this hybrid algorithm, we have adapted the local best and global best factors to align with the exploration task, eliminating the need for a continuous stream of information from the target. All the mathematical formulations in this paper are tailored for a two-dimensional search space, although they can be extended to three-dimensional space for UAV search applications.

### A. Proposed Hybrid Algorithm

The LF movement is well-suited for extensive space exploration missions involving sparse targets due to its inherent characteristics. In order to optimize the paths taken by robots, inter-robot repulsion is taken into consideration, dispersing robots away from each other. The PSO algorithm, which accounts for various behavioral aspects, is employed as a movement type in this hybrid algorithm. The developed hybrid algorithm combines LF movement and the repulsion factor using PSO. PSO computes a weighted average of these two movement components to facilitate effective area exploration. In this context, the local best position of an individual robot corresponds to the position along its path that yields the best fitness value. This concept is adapted to align with the exploration task, where the robot aims to move in the direction and step size derived from the LF method. However, relying solely on the local best position obtained from LF might lead to the robot exploring an area already covered by another robot or currently being explored by one. To mitigate this issue, the global best position is defined as a repulsive force inversely proportional to the distance from other robots, with the direction leading away from them. The pseudocode is shown in Algorithm 1. Further dissemination of optimal positions (local & global) is presented Sections III-A1 and III-A2 respectively.

*1) Local best Position:* The LF represents the optimal method for foraging and various exploration tasks, a strategy employed by numerous species and validated through practical applications. It is assumed that if a single robot is deployed within a search space, LF proves to be a superior exploration strategy when compared to other traditional methods. [34] proposed the integration of LF and an artificial potential field for multi-robot explorations. If the step size from LF is denoted as "$s$" and the movement direction as "$L_\theta$," then the local best position can be determined using the following equation:

$$P_{\text{lb}xj} - x_j = s_j \cos(L_{\theta_j}) \quad (1)$$
$$P_{\text{lb}yj} - y_j = s_j \sin(L_{\theta_j}) \quad (2)$$

---

**Algorithm 1:** Hybrid LF-PSO Algorithm for Multi-Robot Exploration

**Data:** Initial robot positions $(x_i, y_i)$, LF and PSO parameters, other algorithm-specific parameters, maximum iterations $T$

1 **for** $t = 1$ *to* $T$ **do**
2    **foreach** *robot i* **do**
3      Calculate LF step size $s_i$ and direction $L_{\theta_i}$;
4      Update LF position: $(P_{\text{LF}x_i}, P_{\text{LF}y_i}) \leftarrow (x_i + s_i \cdot \cos(L_{\theta_i}), y_i + s_i \cdot \sin(L_{\theta_i}))$;
5      Calculate repulsion force direction $R_{\theta_i}$ based on the positions of other robots;
6      Update repulsion position: $(P_{\text{rep}x_i}, P_{\text{rep}y_i}) \leftarrow (x_i + R_{\theta_i})$;
7      Calculate PSO velocity updates using Equation 6;
8      Update PSO position: $(P_{\text{PSO}x_i}, P_{\text{PSO}y_i}) \leftarrow (x_i + \text{PSO\_velocity})$;
9      Update robot position: $(x_i, y_i) \leftarrow \text{Combine}(P_{\text{LF}x_i}, P_{\text{LF}y_i}, P_{\text{rep}x_i}, P_{\text{rep}y_i}, P_{\text{PSO}x_i}, P_{\text{PSO}y_i})$;
10      $(P_{\text{lb}x_i}, P_{\text{lb}y_i}) \leftarrow \text{UpdateLocalBest}(x_i, y_i, s_i, L_{\theta_i})$;
11      $(P_{\text{gb}x_i}, P_{\text{gb}y_i}) \leftarrow \text{UpdateGlobalBest}(\text{All\_Robots})$;
12    **end foreach**
13 **end for**

**Result:** Final positions of robots and exploration results

---

where, $P_{\text{lb}xj}$ is the local best position in the x-direction and $P_{\text{lb}yj}$ is the local best position in the y-direction, with (x, y) representing the current position of the robot within the cognitive component. Once the robot has traversed a distance equal to the step size, the local best position is subsequently updated based on the new step size and the updated heading direction for the local best term. The pseudocode is shown in Algorithm 2.

---

**Algorithm 2:** Update Local Best Position Using LF

**Data:** Current robot position $(x_j, y_j)$, LF step size $(s_j)$, LF movement direction $(L_{\theta_j})$

1 $P_{\text{lb}xj} - x_j = s_j \cdot \cos(L_{\theta_j})$;
2 $P_{\text{lb}yj} - y_j = s_j \cdot \sin(L_{\theta_j})$;
3 **return** $P_{lbxj}, P_{lbyj}$

**Result:** Updated local best position $(P_{\text{lb}xj}, P_{\text{lb}yj})$

---

*2) Global Best Position and Inter-Robot Repulsion:* The concept of inter-robot repulsion draws inspiration from the methodology presented in [25], where a modified PSO algorithm was employed for exploration purposes. This repulsion factor becomes null in the absence of other robots within its communication range. To facilitate the efficient dispersion of robots and ensure comprehensive coverage during exploration, a novel approach involving robot clusters is introduced. When more than two robots find themselves within communication range, the system defines robot clusters. These clusters are structured in such a way that every robot within the cluster can establish communication with at least one other robot, given a communication range denoted by 'd' as shown in Figure 1.

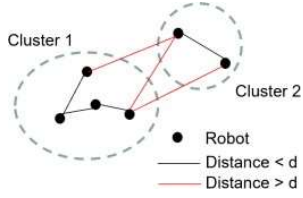

Fig. 1: Virtual clusters established by robots to create a repulsive force between them

Each robot within a cluster exerts a repulsive force on other robots, pushing them away in the direction opposite to the originating robot's position, as depicted in Figure 2. This dispersion mechanism is designed to enable the robots to effectively cover a broad search area. The global best position is anticipated to be located in a direction away from the collective presence of all the robots, and its distance is expected to be the average of the inverses of the distances between the robots within the cluster.

$$D_j = \frac{1}{d_1} + \frac{1}{d_2} \quad (3)$$

$$P_{gbxj} - x_j = D_j \cdot \cos(R_{\theta j}) \quad (4)$$

$$P_{gbyj} - y_j = D_j \cdot \sin(R_{\theta j}) - 11 \quad (5)$$

In Equation (3), $d_1$ and $d_2$ represent the distances of the robot from "Robot 1" and "Robot 2", respectively. In Equation (4), $P_{gbxj}$ and $P_{gbyj}$ denote the global best positions in the x and y directions, respectively. The variables (x, y) represent the current position of the robot in the social component. The modified equation for PSO is shown below:

$$V_{i,j}(t+1) = \omega \cdot V_{i,j}(t) + p_\omega \cdot \text{rand} \cdot (P_{lbj}(t) - x_{i,j}(t))$$
$$+ n_\omega \cdot \text{rand} \cdot (P_{gbj}(t) - x_{i,j}(t)) \quad (6)$$

$$x_{i,j}(t+1) = V_{i,j}(t+1) + x_{i,j}(t) \quad (7)$$

In each time step, each robot adjusts its velocity according to Equation (6). The local best is updated once a robot completes its current LF step size distance, while the global repulsive force is updated when a robot approaches other robots. An individual robot's movement takes into account all three velocity components. In our paper, we intend to fine-tune these parameters to enhance search efficiency. The pseudocode is shown in Algorithm 3.

*B. Obstacle Avoidance*

In contrast to the Robotic PSO [24], which introduces an additional term in the PSO for obstacle avoidance involving both robots and obstacles, we incorporate a distinct policy for

**Algorithm 3:** Hybrid LF-PSO Algorithm with Inter-Robot Repulsion

**Data:** Current robot positions $(x_j, y_j)$, LF step size $(s_j)$, LF movement direction $(L_{\theta j})$, distances from other robots $(d_1, d_2)$

1 **for** t = 1 to T **do**
2    Calculate repulsion force direction $(R_{\theta j})$ based on the positions of other robots;
3    Calculate global repulsion distance $(D_j)$ using $d_1$ and $d_2$;
4    Calculate global best positions updates;
5    **foreach** robot i **do**
6       Calculate LF step size and direction using $s_j$ and $L_{\theta j}$;
7       Calculate velocity updates using PSO (Equation 6);
8       Update robot position based on LF, PSO, and repulsion;
9       Update local best position based on fitness evaluation;
10    **end foreach**
11 **end for**
12 **return** Updated robot positions, $P_{lbxj}$, $P_{lbyj}$, $P_{gbxj}$, $P_{gbyj}$

**Result:** Updated robot positions, local best positions $(P_{lbxj}, P_{lbyj})$, global best positions $(P_{gbxj}, P_{gbyj})$

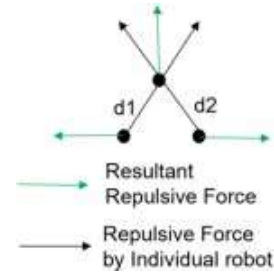

Fig. 2: Direction of repulsion caused by nearby robots

obstacle avoidance in our hybrid algorithm. Since the social component of our hybrid algorithm already considers the influence of neighboring robots, including an extra inter-robot collision avoidance term would be redundant. When our robots approach obstacles other than fellow robots, we employ a straightforward obstacle avoidance policy. The current obstacle avoidance methodology in this paper is adapted from the approach outlined in [26]. As a robot traverses the search space following the hybrid algorithm, a conditional statement monitors obstacle collision avoidance based on sensor data as shown in Table I. The obstacle avoidance algorithm activates as soon as at least one sensor is triggered, and it remains active until the robot safely clears the obstacle. During this obstacle avoidance phase, the robot temporarily departs from the rules of the hybrid algorithm. Each robot is assumed to be equipped with three sensors positioned at the front, left, and right sides to detect obstacles when they are at a certain distance. When a robot detects other robots or obstacles, it adheres to specific movement rules to avoid collisions.

TABLE I: Obstacle Avoidance Policy

| Moving Action | Left Sensor | Front Sensor | Right Sensor |
|---|---|---|---|
| Turn $\frac{\pi}{4}$ Left/Right | FALSE | TRUE | FALSE |
| Turn $\frac{\pi}{2}$ Left | FALSE | TRUE | TRUE |
| Turn $\frac{\pi}{2}$ Right | TRUE | TRUE | FALSE |
| Turn $\pi$ | TRUE | TRUE | TRUE |
| Turn $\frac{\pi}{2}$ Right | TRUE | FALSE | FALSE |
| Turn $\frac{\pi}{2}$ Left | FALSE | FALSE | TRUE |

*C. Path Representation and Encoding*

In the context of the multi-robot exploration problem, the representation of paths for individual robots and their encoding play a crucial role in the effectiveness of the hybrid LF-PSO algorithm. To facilitate comprehensive area coverage, each robot's path is represented as a sequence of waypoints in the two-dimensional search space.

**Waypoint Representation:** A waypoint serves as a critical reference point for a robot's movement. It defines a specific location in the search space and acts as an anchor point for the robot to navigate towards. The waypoints are strategically positioned to ensure efficient exploration while avoiding overlap with other robots' paths. The waypoints are encoded with the following attributes: **(1) Position:** Each waypoint is associated with a set of coordinates $(x, y)$, which represent its location in the search space. These coordinates determine where the robot should move during its exploration. **(2) Connectivity:** Waypoints are interconnected to form a path that guides the robot through the search area. The order in which waypoints are connected determines the robot's trajectory. **(3) Exploration Status:** Each waypoint maintains an exploration status to indicate whether it has been visited by the robot or remains unexplored. This status is crucial for the algorithm to track the progress of the exploration.

**Path Encoding:** The paths for individual robots are encoded as a series of waypoints. The encoding format is structured to ensure smooth navigation and efficient coverage of the search space. A robot's path can be represented as an ordered list of waypoints, starting from an initial position and concluding when the exploration mission is complete. For example, consider a robot starting at position $(x_0, y_0)$. Its path encoding might look like this:

$$Path_{\text{Robot 1}} = [(x_0, y_0), (x_1, y_1), (x_2, y_2), \ldots] \quad (8)$$

In this representation, the robot's path is defined by a sequence of waypoints, with each waypoint containing the necessary information for the robot to navigate effectively. The LF-PSO algorithm operates by adjusting the positions of these waypoints based on LF movement, PSO optimization, and inter-robot repulsion to achieve efficient exploration. The hybrid LF-PSO algorithm described in Algorithm 1 optimizes the positions of these waypoints to guide the robots through the search space while balancing exploration and repulsion factors. This encoding ensures that the robots' paths are adaptable and responsive to both local and global exploration objectives, ultimately leading to a comprehensive coverage of the unfamiliar environment.

## IV. EXPERIMENTAL RESULTS

To assess the performance of the proposed algorithm, we conducted simulations of the hybrid algorithm within a two-dimensional Python environment. The simulation took place in a 20m * 20m area devoid of obstacles. The robots were assumed to possess a communication range of 2m and a maximum speed of 0.2m/s. The beta term for LF was set to 1, and other PSO parameters, specifically $\omega$, $p_\omega$, and $n_\omega$, were assigned values of 0.6, 2, and 2, respectively. In this simulation, each iteration was equivalent to a one-second timestep. The robot's size was considered as 0.6 * 0.6m, and the area covered by a robot based on its size was treated as explored. Initially, there were no targets in the search space, and the performance metric was simply the amount of area covered within the total search space volume. For comparison, we selected the traditional PSO, and LF since it is suitable for the chosen search scenario to compare with the non-hybrid versions of the hybrid algorithm. We conducted 20 trial simulations, each involving 10 robots within the environment. The results presented are the averages obtained from these 20 simulations for each test scenario.

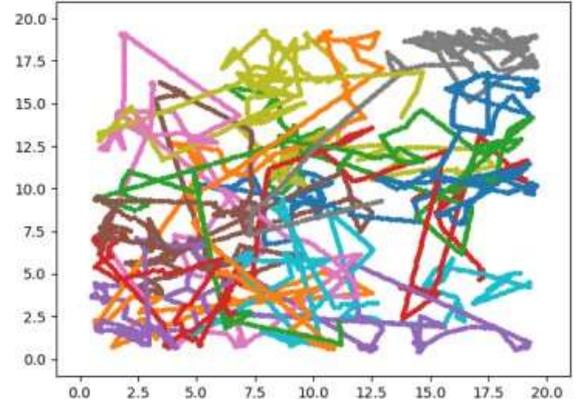

Fig. 3: Simulation of the paths taken by 10 robots as they initiate from random positions within a 20m * 20m exploration area, spanning a duration of 600 time steps via hybrid algorithm

When the robots were initially placed randomly within the search space, the hybrid algorithm covered an average of 76.49% of the exploration area in 600 timesteps across the 20 simulations as shown in Figure 3. The hybrid algorithm achieved the highest exploration efficiency in this scenario due to its ability to incorporate both PSO and LF components. PSO helps optimize the robot movement based on global information, while LF introduces repulsion to avoid clustering, ensuring better coverage. Figure 4, on the other hand, presents an example of one simulation result using the PSO method for 10 robots starting at the same random locations, with an area coverage of 61.13%. In contrast, the PSO achieved approximately 61.13% coverage but couldn't match the hybrid's performance. It relies solely on particle, which might lead to suboptimal local minima. Similarly, Figure 5 shows simulation result using the LF method. In which, LF achieved 58% and being a local repulsion-based method, it had a decent exploration efficiency but struggled to disperse robots efficiently compared to the hybrid algorithm.

Figure 6 showcases an example of the simulation results using the hybrid algorithm with 10 robots starting at the corner of the search space at coordinates (10, 2), achieving an area coverage of 73.15% over 600 timesteps. Even in this challenging scenario, the hybrid algorithm outperformed the other methods. It maintained a higher exploration efficiency due to the repulsion component, which prevented robots from getting stuck in one corner. In contrast, for the PSO method, the average area coverage dropped to around 52% when averaged over the 20 simulation runs as shown in Figure 3. PSO's performance dropped significantly as all robots starting from one corner made it challenging to optimize their movements collectively. Whereas, LF achieved 51%, with robots struggling to disperse efficiently as shown in Figure 8.

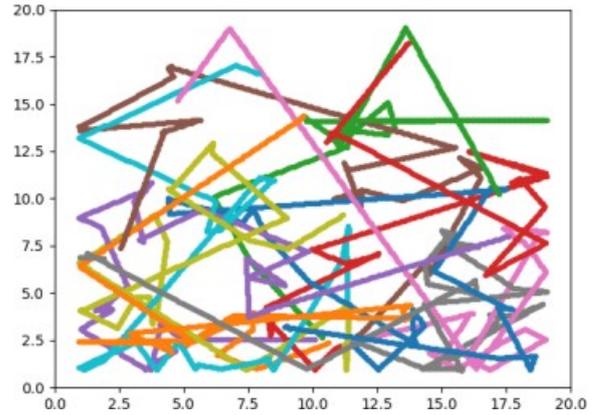

Fig. 5: Simulation of the paths taken by 10 robots as they initiate from random positions within a 20m * 20m exploration area, spanning a duration of 600 time steps via LF method

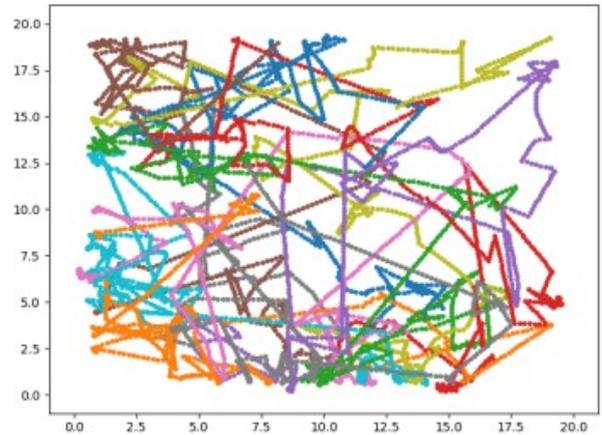

Fig. 6: Simulating the movement paths of 10 robots that commence their journey from the coordinates (10, 2) within a 20m * 20m exploration area, spanning a duration of 600-time steps via hybrid algorithm

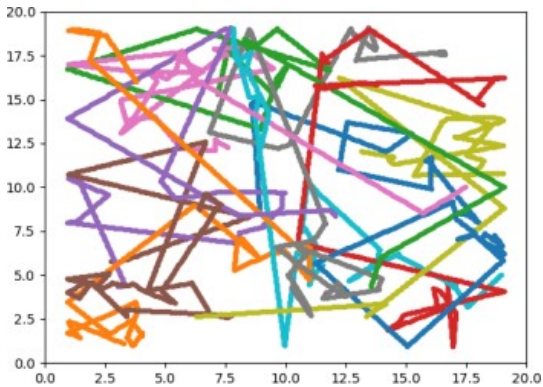

Fig. 4: Simulation of the paths taken by 10 robots as they initiate from random positions within a 20m * 20m exploration area, spanning a duration of 600 time steps via PSO method

### A. Additional Experiments

The simulations were conducted in a controlled two-dimensional Python environment, offering insights into the algorithm's effectiveness and highlighting the algorithm's promise in exploration tasks, both in terms of efficiency and its ability to adapt to varying scenarios as shown in Tables II and III.

**Experiment 1 (Varying Robot and Victim Counts):** In the "Small Scenario," which featured a lower number of robots exploring a victim-less environment, the algorithm demonstrated exceptional exploration efficiency, covering an average of 76% of the exploration area within 600 time steps. Although no victims were present in this scenario, it's noteworthy that the algorithm efficiently explored the environment while encountering 2 collisions. Additionally, it consumed 5.4 kWh of energy, reflecting the computational load and movement during exploration. Transitioning to the "Large Scenario," where more robots were employed in the exploration, the algorithm maintained commendable performance. It achieved 72% exploration efficiency within the same timeframe. Interestingly, in this case, 2 victims were found, showcasing the algorithm's potential for victim discovery even in victim-sparse scenarios. This scenario also encountered 1

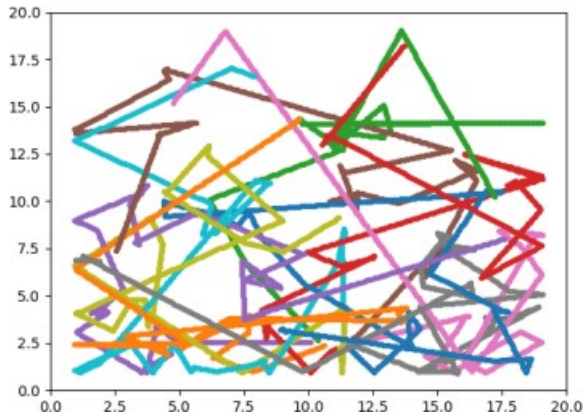

Fig. 7: Simulating the movement paths of 10 robots that commence their journey from the coordinates (10, 2) within a 20m * 20m exploration area, spanning a duration of 600 time steps via PSO method

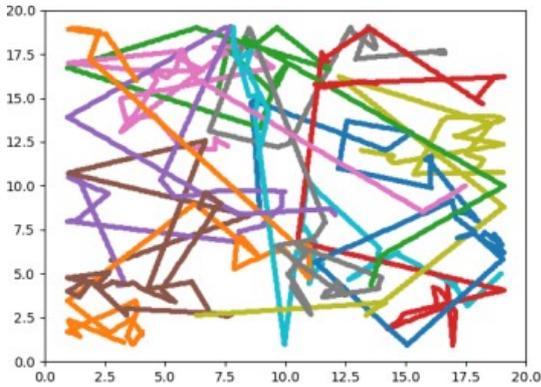

Fig. 8: Simulating the movement paths of 10 robots that commence their journey from the coordinates (10, 2) within a 20m * 20m exploration area, spanning a duration of 600 time steps via LF method

collision, and energy consumption increased slightly to 7.1 kWh due to the larger number of robots involved.

**Experiment 2 (Obstacle Density and Complexity):** In scenarios with varying obstacle density and complexity, the algorithm's robustness became evident. In the "Low Complexity" scenario, characterized by few or no obstacles, the algorithm excelled with an exploration efficiency of 76%. Despite the absence of victims, it performed efficiently, and no collisions occurred, emphasizing its adaptability to open spaces. Energy consumption in this scenario reached 6.2 kWh, driven by computational tasks and movement. In contrast, the "High Complexity" scenario, marked by intricate environments with numerous obstacles, maintained strong performance. The algorithm achieved 72% exploration efficiency, showcasing its ability to navigate complex spaces. Importantly, in this scenario, the algorithm discovered 3 victims, emphasizing its capability for victim detection even in challenging environments. It encountered 2 collisions due to the increased complexity, and energy consumption rose to 8.0 kWh.

TABLE II: Experiment 1: Varying Robot and Victim Counts

| Scenario | Small Scenario | Large Scenario |
|---|---|---|
| Exploration Efficiency (%) | 76% | 72% |
| Time Taken (s) | 600 | 600 |
| Victims Found | 0 | 2 |
| Collision Count | 2 | 1 |
| Energy Consumption (kWh) | 5.4 | 7.1 |

TABLE III: Experiment 2: Obstacle Density and Complexity (Low & High)

| Scenario | Complexity | Complexity |
|---|---|---|
| Exploration Efficiency (%) | 76% | 72% |
| Time Taken (s) | 600 | 600 |
| Victims Found | 1 | 3 |
| Collision Count | 0 | 2 |
| Energy Consumption (kWh) | 6.2 | 8.0 |

## V. CONCLUSION

In conclusion, this paper introduces a novel hybrid algorithm that combines LF and PSO for efficient multi-robot exploration in unknown environments. The significance of this research lies in its applicability to real-world scenarios where robots must operate with limited information, communication, and in GPS-denied environments. Our hybrid algorithm leverages LF's ability to cover large areas with sparse targets and combines it with the inter-robot repulsion concept from PSO. By doing so, it offers a solution to the challenge of exploring unknown areas without prior knowledge of the search space's size or obstacles. The proposed algorithm addresses several common assumptions made in previous research, including the lack of information about the search area, robot positions, communication restrictions, and limited information about the target's position. By doing this, it offers a more practical approach to multi-robot exploration.

Preliminary simulation results demonstrate the effectiveness of the hybrid algorithm in terms of area coverage compared to existing methods. When robots are randomly placed in the search space, our algorithm covers a significant portion of the area, outperforming traditional random walk methods. Even when robots start in one corner of the room, the algorithm excels at dispersing robots and exploring a wide area. Future work will involve refining obstacle avoidance policies, introducing obstacles to the environment, and testing the algorithm with target detection tasks. Additionally, parameter tuning and further simulations will be conducted to optimize the algorithm's efficiency. In summary, our hybrid LF-PSO algorithm presents a promising approach to coordinated multi-robot exploration in challenging real-world scenarios, offering a potential solution for tasks such as USAR operations and so on. Its adaptability and performance make it a valuable addition to the field of robotics and autonomous systems.